\documentclass[conference]{IEEEtran}
\IEEEoverridecommandlockouts
\usepackage{cite}
\usepackage{amsmath,amssymb,amsfonts}
\usepackage{algorithmic}
\usepackage{graphicx}
\usepackage{subfigure}
\usepackage{textcomp}
\usepackage{xcolor}
\usepackage{booktabs}
\usepackage{amssymb}
\usepackage{threeparttable}
\usepackage{diagbox}
\usepackage{makecell}
\usepackage{multirow}
\usepackage{tabularx}
\usepackage{afterpage}
\def\BibTeX{{\rm B\kern-.05em{\sc i\kern-.025em b}\kern-.08em
    T\kern-.1667em\lower.7ex\hbox{E}\kern-.125emX}}
\DeclareRobustCommand*{\IEEEauthorrefmark}[1]{%
	\raisebox{0pt}[0pt][0pt]{\textsuperscript{\footnotesize\ensuremath{#1}}}}
\begin{document}

\title{IAFI-FCOS: Intra- and across-layer feature interaction FCOS model for lesion detection of CT images
}

\author{
	\IEEEauthorblockN{
		Qiu Guan\IEEEauthorrefmark{1},
		Mengjie Pan\IEEEauthorrefmark{1},
		Feng Chen\IEEEauthorrefmark{2},
		Zhiqiang Yang\IEEEauthorrefmark{1},
		Zhongwen Yu\IEEEauthorrefmark{1},
		\\Qianwei Zhou\IEEEauthorrefmark{1},
		Haigen Hu\IEEEauthorrefmark{1}\IEEEauthorrefmark{*} \thanks{* Corresponding author: Haigen Hu}} 
	\IEEEauthorblockA{\IEEEauthorrefmark{1}\textit{ College of Computer Science and Technology, Zhejiang University of Technology, Hangzhou, China}}
	\IEEEauthorblockA{\IEEEauthorrefmark{2}\textit{ Department of Radiology, The First Affiliated Hospital, Zhejiang University School of Medicine, Hangzhou, China} \\
	\{gq, 211122120107, zqw, hghu\}@zjut.edu.cn, \\
	chenfenghz@zju.edu.cn, 874125760@qq.com, vincentyu67373@gmail.com}
}
\maketitle

\begin{abstract}
Effective lesion detection in medical image is not only rely on the features of lesion region, but also deeply relative to the surrounding information. However, most current methods have not fully utilize it. What’s more, multi-scale feature fusion mechanism of most traditional detectors are unable to transmit  detail information without loss, which makes it hard to detect small and boundary-ambiguous lesion in early stage disease. To address the above issues, we propose a novel intra- and across-layer feature interaction FCOS model (IAFI-FCOS) with a multi-scale feature fusion mechanism ICAF-FPN, which is a network structure with intra-layer context augmentation (ICA) block and across-layer feature weighting (AFW) block. Therefore, the traditional FCOS detector is optimized by enriching the feature representation from two perspectives. Specifically, the ICA block utilizes dilated attention to augment the context information in order to capture long-range dependencies between the lesion region and the surrounding. The AFW block utilizes dual-axis attention mechanism and weighting operation to obtain the efficient across-layer interaction features, enhancing the representation of detailed features. Our approach has been extensively experimented on both the private pancreatic lesion dataset and the public DeepLesion dataset, with AP\textsubscript{50} of 62.2\% and 60.0\%, respectively, and these results are 6.4\% and 2.3\% higher than the FCOS. Additionally, our model achieves SOTA results on the pancreatic lesion dataset.
\end{abstract}


\section{Introduction}

Cancer is a major global public health concern, with 10 million people worldwide succumbing to cancer by the year 2020. The chances of survival would significantly increase if cancer is detected early \cite{b1}. However, there is currently a lack of simple and efficient lesion detection methods, resulting in the discovery of most cancers at later stages. For example, pancreatic Ductal Adenocarcinoma (PDAC) is often treated only after the appearance of metastatic symptoms, which leads to an increase in mortality rate \cite{b2}.

In the early stages of disease diagnosis, radiologists would normally screen for tumors using medical imaging, such as Computed Tomography (CT). However, the accuracy of diagnose is often relies on the experience of the medical professionals, and when screening a large number of CT scan images, it can consume a considerable amount of their time and energy, leading to the possibility of errors and omissions \cite{b3}. Utilizing automated Computer-Aided Diagnosis (CAD) systems can assist doctors automatically identifying suspected lesion locations and reduce the workload for physician, enabling faster and more accurate early cancer screening.

\begin{figure}
	\centering
	\subfigure[easy to detect]{
		\begin{minipage}[b]{0.2\textwidth}
			\includegraphics[width=1\linewidth]{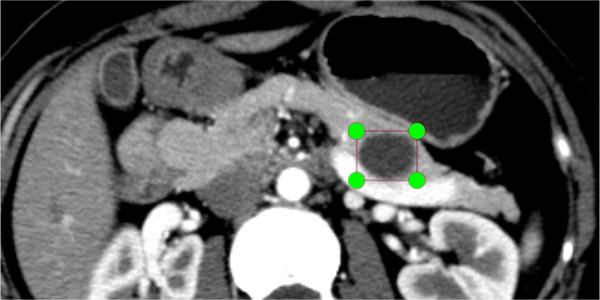}\vspace{2pt}
			\includegraphics[width=1\linewidth]{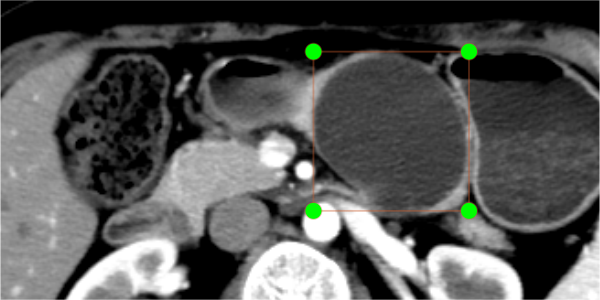}\vspace{4pt}
	\end{minipage}}
	\subfigure[hard to detect]{
		\begin{minipage}[b]{0.2\textwidth}
			\includegraphics[width=1\linewidth]{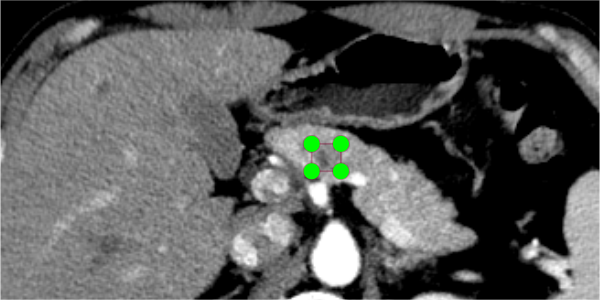}\vspace{2pt}
			\includegraphics[width=1\linewidth]{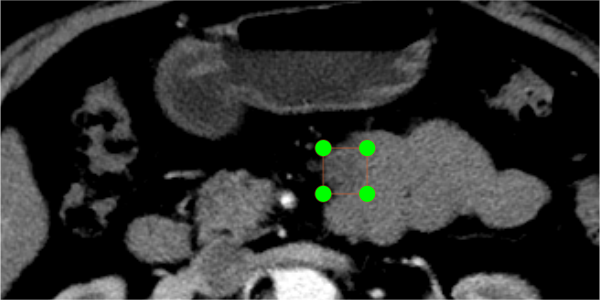}\vspace{4pt}
	\end{minipage}}
	\caption{CT scan visualization of the pancreas dataset. (a) easy to detect: tumor features are distinct and have clear boundaries. (b) hard to detect: tumor boundaries are fuzzy and the target is small, a situation that is often difficult for the network to identify.}
	\label{figure 1}
\end{figure}

Computer-Aided Detection (CADe) is a component of CAD, which is aim to detect lesion areas by object detection methods. Directly applying these methods to lesion detection in CT scan images may not guarantee the satisfactory performance since there's a difference between medical images and natural images. To the best knowledges we know, there exist several challenges as follows:

Firstly, there is often a correlation between the type of lesion and its location. Such as mucinous cystic neoplasms (MCNs), a type of pancreatic cystic neoplasms (PCNs), most commonly occurs in the pancreatic body and tail \cite{b5}. Existing methods \cite{b29} mainly focus on the area of the lesion and do not fully utilize the information surrounding the lesion. Secondly, the traditional detectors \cite{b8,b10,b17} has two drawbacks when performing multi-scale feature fusion: (1) direct fuse of different layers of features reduces the representation of multi-scale features, (2) top-down transfer of features leads to loss of information. When the lesion area occupies only a small fraction of the pixels in CT scan image, or when the difference in features between the lesion area and the non-lesion area is not obvious, the loss of detail information makes it difficult to identify and localize the lesion. Fig.~\ref{figure 1} illustrate visualization of CT scan with distinctive and less distinctive lesion features. 

To address the above challenges, this paper proposes a novel intra- and across-layer feature interaction FCOS model (IAFI-FCOS). The main design a multi-scale feature fusion mechanism ICAF-FPN with intra-layer context augmentation (ICA) block and across-layer feature weighting (AFW) block. This model enriches the lesion-related features and improves the accuracy of early cancer detection. The main contributions of this paper are as follows:
\begin{itemize}
\item The proposed ICA block, at each intra-layer, utilizes dilated attention transformer to increase the receptive field, supplement contextual information in the lesion area, and learn long-range dependencies with surrounding.
\item The proposed AFW block utilizes dual-axis attention to aggregate across-layer features, then the aggregated features adaptively filter redundant and conflicting information through weighting when fused with features of each layer. This complement the texture information and position information of small targets from low-level feature maps, alleviating the issue of information loss from traditional methods. 
\item The proposed IAFI-FCOS model is train and validate on both the private pancreatic lesion dataset and the public DeepLesion dataset, results in better performance compared to the other methods, which proves its performance and its robustness on different datasets.
\end{itemize}

The rest of the paper is organized as follows. We provide an overview of object detection methods and its applications on the field of medical imaging in Section \uppercase\expandafter{\romannumeral2}. In Section \uppercase\expandafter{\romannumeral3}, the method proposed in this paper is specifically described. In Section \uppercase\expandafter{\romannumeral4}, the results are reported and analyzed. Finally, conclusions are drawn in Section \uppercase\expandafter{\romannumeral5}.
\section{Related Work}

\textbf{Object Detectors}. Object detection is a fundamental task in the field of computer vision, and numerous research achievements have propelled its advancement. The object detectors are generally classified into three categories: two-stage detectors \cite{b8, b9, b10, b11, b12}, one-stage detectors \cite{b14, b15, b16, b17, b18, b19, b20} and transformer-based detectors \cite{b21, b22, b23}. The two-stage detectors follow the traditional object detection pipeline, by generating lots of candidate regions and then classifying the objects present in each candidate box into different object classes. The one-stage detectors operate as either a regression or classification problem, directly mapping from image pixels to bounding box coordinates and class confidences. Transformer-based detectors are end-to-end structures, eliminating the need for manually designed components (e.g., non-maximum suppression). 

For the lesion detection task, we choose the one-stage detector FCOS \cite{b17} as our baseline model. This is because its simplifies the whole target detection process and performs better when dealing with small targets.

\textbf{Feature Pyramid Networks}. Object detection tasks commonly utilize Feature Pyramid Networks (FPN) \cite{b10} to enable models to effectively detect objects at different scales. SSD \cite{b14} first attempts to predict the location and class of targets with multi-scale features. FPN \cite{b10} introduces a top-down and lateral connection mechanism, effectively fusing features at different scales. Subsequently, PANet \cite{b24} further proposes a bottom-up path, which combined with FPN to enable high-level features to capture detailed information in low-level features. NAS-FPN \cite{b25} optimizes the Feature Pyramid Network through neural architecture search to achieve automatic search and design and enhance the performance of object detection models. FCOS uses the traditional FPN structure with simple top-down fusion of multi-scale features, which can lead to information loss.

\textbf{Medical Images lesion Detection}. Medical image lesion detection, which is used of computer-aided detection (CADe) to identify the location and category of lesions [9]. With the help of CADe, the time and computational cost required can be reduced, assisting physicians to improve efficiency. For example, Ding et al. \cite{b26} and Zhu et al. \cite{b27} improve the Faster RCNN \cite{b8} and combine it with 3D convolution for the detection of lung nodules, which enhance the accuracy of nodule identification by strengthing improve the fine-grained representation and capturing more unique features. Fan et al. \cite{b28} develops a framework for Computer-Aided Diagnosis (CAD) system based on Mask Region Convolutional Neural Network \cite{b11} for the large-scale detection and segmentation of breast cancer. These frameworks all utilize 3D CNN to enhance the two-stage detector, improving detection performance, but they require substantial computational resources and extensive manual annotation of 3D bounding boxes. Subsequently, Liu et al. \cite{b29} further explores the one-stage method YOLOv3 to improve the detection performance in universal lesion areas. It significantly strengthened the accuracy of the lesion detector using only a 2D structure. However, there still remain some challenges to recognize those tough lesions, such as the smaller lesions and those with ambiguous borders.

\begin{figure*}
	\centering
	\includegraphics[width=0.95\linewidth]{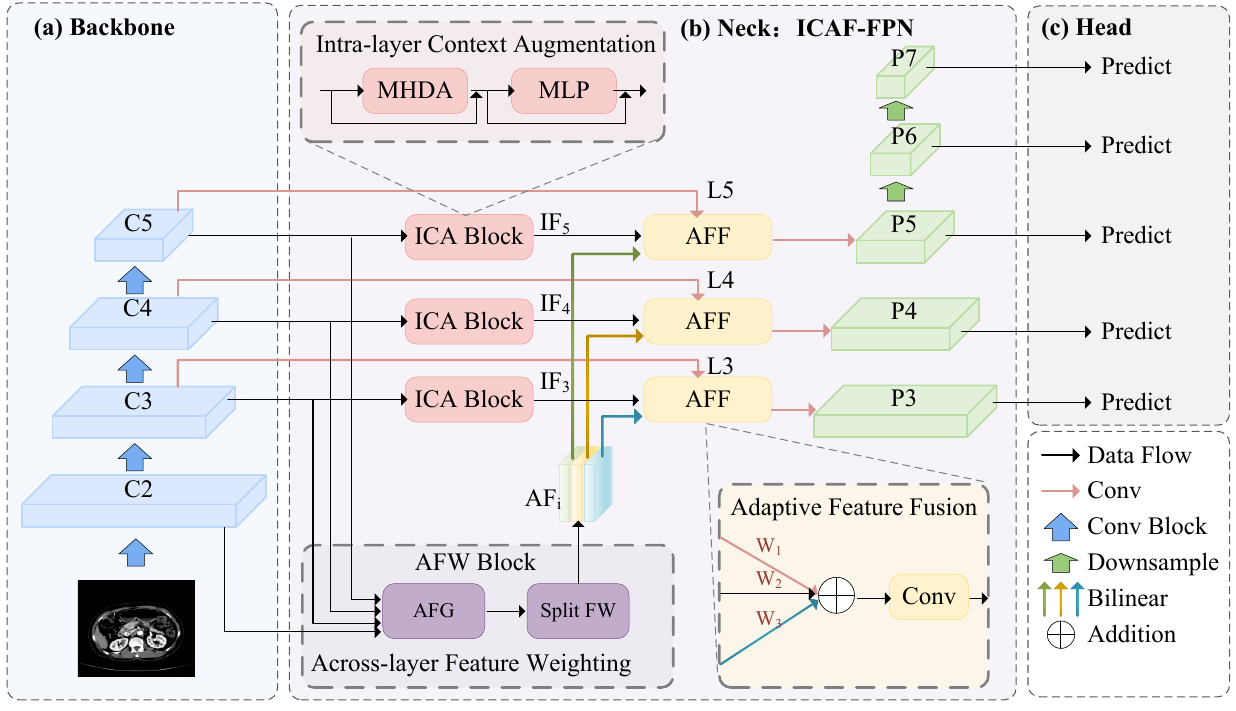}
	\caption{Overview of the network architecture of the IAFI-FCOS detection framework, which mainly consists of three components: (a)a backbone network for feature extraction, (b)the ICAF-FPN neck and (c)the object detection head network. The ${C_{i}}$, ${P_{i}}$, ${IF_{i}}$, ${AF_{i}}$ and ${L_{i}}$ represent feature maps, the ${W_{i}}$ indicate learnable weights. } 
	\label{figure 2}
\end{figure*}

To address above issues, we propose a novel IAFI-FCOS method to capture effective information for CT scan images, improving the accuracy and robustness of lesion detection. 

\section{Method}
In this section, we present the details of the proposed the IAFI-FCOS. This study utilizes the FCOS as our baseline, and combines it with a novel proposed Neck called ICAF-FPN. The overall architecture of the network is illustrated in Fig.~\ref{figure 2}. It mainly consists of the following components: (a) Backbone: Given an input feature map I, extract multi-scale features ${C_{i} (i = 2, 3, 4, 5)}$ using convolution network, i.e. ResNet. (b) Neck: This part consists of ICAF-FPN. The main role is to aggregate and distribute the multi-scale features ${C_{i}}$ from the backbone network. (c) Head: Outputs the final classification and localization prediction results.

\subsection{ICAF-FPN}
The accuracy of object detection relies on the processing of features at different scales by the NECK part. Low-level features tend to carry texture information and edge information, which helps in the localization of small targets. High-level features include semantic features and the location of larger objects. Effectively fuse features of different levels can improve the network's detection accuracy for objects of different sizes \cite{b30}. For the lesion detection task, we design a new multi-scale feature fusion mechanism, ICAF-FPN, which prevents information loss by aggregating features in both intra- and across-layer perspective.

\textbf{Intra-layer.} The recognition of a lesion depends not only on its inherent feature information, but also needs to be aided by the information provided by the background surroundings nearby the lesion. Traditional approaches employ convolution to process features at each layer, to capture the local features of the target. However, the nature of the convolutional operation, i.e., the limited receptive field, unavoidably has some drawbacks in establishing global dependencies. In our work, the ICA block is designed to deeply interact with each layer's inner features. We utilize multi-head dilated attention (MHDA) block to capture the global dependencies between the lesion area and the surrounding pixels, which significantly expands the receptive fields. This approach preserves the model's sensitivity to local features while capturing global contextual information, enhancing the accuracy of lesion identification. 

\textbf{Across-layer.} When inconspicuous lesion boundaries and small lesions appear (such as in Fig.~\ref{figure 1} (b)), the amount of detail information determines the success of screening and detection. However, the traditional top-down and bottom-up transmission modes may lead to information loss, as each layer only receives complementary neighboring information while the cross-layer information (e.g. ${C_{2}}$ to  ${C_{4}}$) is weakened and lost during transmission. Inspired by Glod-YOLO \cite{b30}, we propose the AFW block, which first extract global effective features directly through across-layer feature gather (AFG) block to reduce the information loss. To retain more information about the positions of small targets, we specifically introduce the low-level feature map ${C_{2}}$ into the AFG block. After extracting the global features, they are assigned to different hierarchical levels. It's worth noting that directly fusing information of different densities may lead to semantic conflicts, thus limiting the expression of multi-scale features. Here, we design the split feature weighting (Split FW) block with different weights for each levels to fuse global features and filter conflicting information.

Finally, the layer features, the intra-layers interaction features and the across-layer interaction features are adaptively fused to obtain the output features of the neck by adaptive feature fusion (AFF) mechanism. Compared with existing FPN, our proposed ICAF-FPN is more focused on the lesion detection task, which not only establishes global dependencies within each layer, but also realizes the depth of the information interaction between the different layers, further improves the expression ability of multi-scale features. This design alleviates the problems in lesion detection.
\subsection{Intra-layer Context Augmentation Block}

\begin{figure}[htb]
	\centering
	\includegraphics[width=0.95\linewidth]{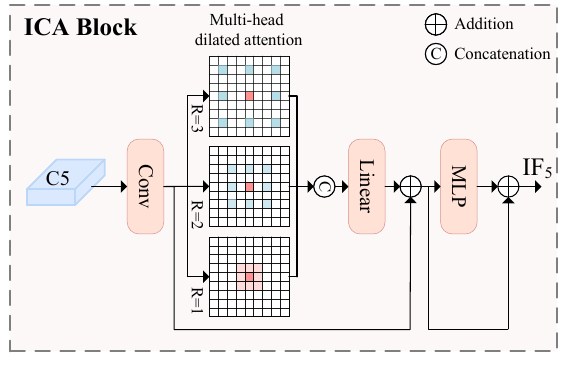}
	\caption{The structure of ICA block.}
	\label{figure 3}
\end{figure}

Dilated convolution \cite{b31} and DilateFormer \cite{b32} expand the receptive field and capture rich contextual information by
setting different dilation rates. To establish the relationship
between the lesion region and the surrounding background,
we design the ICA block. The structure is shown in Fig.~\ref{figure 3}, for the feature maps ${C_{i}(i = 3, 4, 5)}$, we first apply convolution to enrich local details. Subsequently, we utilize a multi-head dilated attention (MHDA) to establish dependencies between pixels. The feature map is divided into three heads by channel dimension and compute the dilated attention for each head. The dilated attention expands the receptive field by setting different the dilation rate R (e.g., R=1, 2, 3). Such as, an unflod operation utilizes a ${3\times3}$ kernel size with R = 3, and the size of  receptive field is ${7\times7}$. Then, we concatenate the features from the different receptive field of three heads together and feed the concatenated features into a linear layer. The residual structure 
complement detailed information. Finally, the intra-layer interaction features ${IF_{i}}$ is obtained through MLP and residual layer.

The above processes can be formulated as:
\begin{equation}
	h_{n}=MHDA(C_{i}^{'}, R_{n}),  \ \  1 \leq n \leq 3,
\end{equation}
\begin{equation}
	X_{i}=C_{i}^{'} + Linear(Concat[h_{1}, h_{2}, h_{3}]),
\end{equation}
\begin{equation}
	IF_{i}=X_{i}+MLP(X_{i}),
\end{equation}
where ${C_{i}^{'}}$ is the feature map after convolution, ${R_{n}}$ represent the dialation rate of the n-th head.

In contrast to conventional multi-head attention which calculate self-attention on the whole graph, we capture the spatial positional connections by building the different receptive fields sparsely. This approach successfully reduces computational complexity while captures global dependency relationships.

\subsection{Across-layer Feature Weighting Block}

\begin{figure}[htb]
	\centering
	\includegraphics[width=0.95\linewidth]{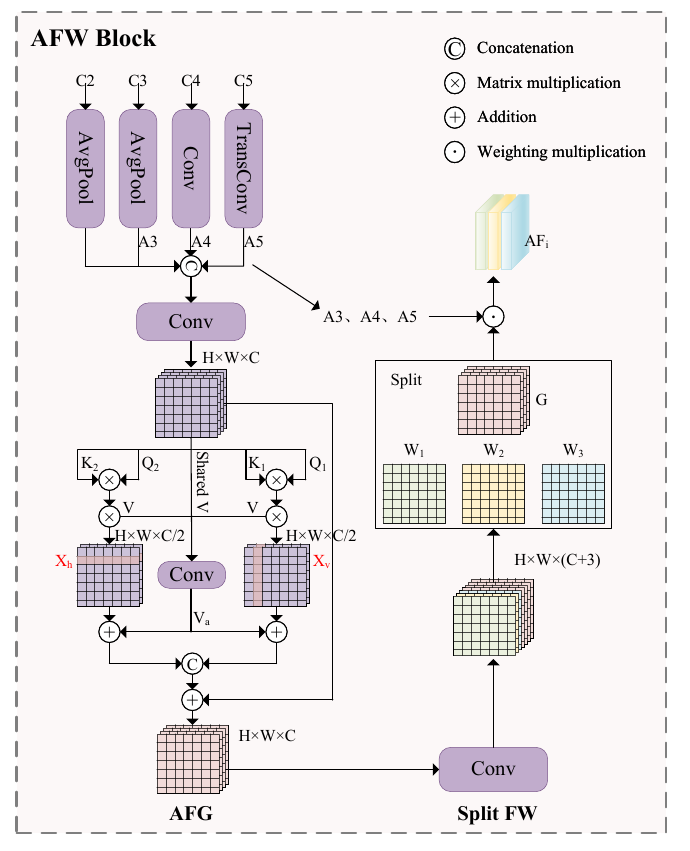}
	\caption{The structure of AFW block. The left and right sides of the figure  represent the specific processes of AFG and Split FW, respectively.}
	\label{figure 4}
\end{figure}

The AFW block consists of the across-layer feature gather (AFG) block and the split feature weighting (Split FW) block, as illustrated in Fig.~\ref{figure 4}. Firstly, the AFG block aligns multi-layer features and extracts global features using a dual-axis attention operation \cite{b33}. Then, the cross-layer global features are weighted and fused with the aligned features, effectively facilitating the interaction of inter-layer information.

\textbf{AFG.} In the neck part of FCOS, a top-down mechanism is directly adopted to merge the ${C3}$, ${C4}$ and ${C5}$. To preserve more detailed information for smaller targets, we introduce features from the C2 layer and employ a different fusion mechanism. Initially, We align the four layers of different scale features to a unified size, avoiding computational overload by mapping images of different resolutions to the size of the C4. For high-resolution feature maps, the average pooling operation is used for downsampling, while the transpose convolution is taken for upsampling the low-resolution images. Due to the reason that the traditional up-sampling methods may lead to loss of information, we adopt the transposed convolution allows flexibility in retaining and reconstructing the information in the original input by learning the parameters. The formula is as follows:
\begin{equation}
	A_i = F_{align}(C_i), \ \ i=2,3,4,5
\end{equation}
where ${F_{aling}}$ represent the alignment method, ${A_{i}}$ represents the aligned features.

The aligned features are concatenated through the channel dimension and the global features are extracted by fusing the spliced features using the dual-axial attention mechanism. The dual-axial attention refer to the establishment of long-range dependencies in the vertical and horizontal directions, respectively. The feature map X is divided into two parts ${X_{v}\in {R^{H\times W \times C/2}}}$ and ${X_{h}\in {R^{H\times W \times C/2}}}$ by channel dimension. In the vertical direction, ${X_v}$ is evenly split into W non-overlapping
vertical axial stripes and projected as ${Q_{1}}$, ${K_{1}}$. In the horizontal direction, ${X_h}$ is evenly split into H non-overlapping horizontal axial stripes and projected as ${Q_{2}}$, ${K_{2}}$. ${V}$ is projected by the feature map ${X}$ and shared in dual-axial. To compute self-attention separately, the formula is as follows:
\begin{equation}
	X_{v}^{'} = Attention(Q_{1}, K_{1}, V)
\end{equation}
\begin{equation}
	X_{h}^{'} = Attention(Q_{2}, K_{2}, V)
\end{equation}
where ${X_{v}^{'}}$ and ${X_{v}^{'}}$ denote the feature map after self-attention.

While the dual axial attention mechanism reduces the complexity of attention, the axis to axis interaction information is lost which is very critical for object detection task. Hence, convolutional are employed to spatially interact with the shared ${V}$, supplementing the connections between different axis. The final output of two parts are concatenated along the channel dimension. The process is formulated as:
\begin{equation}
	X^{'} = X + Concat(X_{v}^{'}+V_{a}, X_{h}^{'}+V_{a})
\end{equation}
where ${V_{a}}$ denotes the value after interacting with the axial information, ${X^{'}}$ denotes the output of AFG.

\textbf{Split FW.} The AFG effectively aggregates across-layer information. In order to efficiently fuse global information into different layers, we use different weights to enhance features at each scale. We expand the feature map ${X^{'}}$ along the channel dimension by adding three dimensions, serving as learnable weights for each layer. After utilize the Split operation, these weights are divided into a globally effective feature weight map ${G}$ and layer-specific feature weight maps ${W_{i} (i=1, 2, 3)}$. ${W_{i}}$ and ${G}$ are multiplied by the aligned feature map ${A_{i}}$, respectively. It can complement the inter-layer correlated features while suppressing the conflicting information to enhance the representation of multi-scale features. The above processes can be formulated as:
\begin{equation}
	AF_{i} = A_{i}\cdot G \cdot W_{i}, \ \ i=3, 4, 5
\end{equation}

\subsection{Adaptive Feature Fusion}
The ICA and AFW blocks aggregate intra- and acoss-scale features, with the ultimate goal of fusing multiple semantic features: ${IF_{i}}$, ${AF_{i}}$ and ${L_{i}}$. ${L_{i}}$ is obtained by 1×1 convolution of ${C_{i}}$. We observe that a straightforward addition leads to confusion of distinct features, which reduce the ability to recognize the target. To address this problem, we employ an adaptive fusion mechanism, introducing weighting parameters to ensure that the contribution of each feature to the final fusion result is learned by the network. This process can be expressed as:
\begin{equation}
	P_{i} = \alpha_{1} \cdot IF_{i} + \alpha_{2} \cdot AF_{i} + \alpha_{3} \cdot L{i}, \  \alpha_{1}+\alpha_{2}+\alpha_{3}=1
\end{equation}
where ${\alpha_{i}}$ denotes different weights, ${P_{i}}$ denotes the output of the neck part.

\section{Materials and Experiment}
In this section, we introduce the datasets, the training schedule and the evaluation metrics used in our experiments. Then, we compare our proposed method with the other methods and analyze the effectiveness of our methodology by ablation study. Patient data were fully anonymized in this study to ensure confidentiality and privacy.

\begin{table*}[htbp]
	\caption{Comparison of detection performance of different algorithms on the pancreas dataset. Use of two evaluation metrics: AP and FROC. The backbone part remains consistent, utilizing ResNet50.}
	\begin{center}
		\begin{tabular}{cccccccccccc}
			\toprule 
			\multirow{2}{*}{Method} & \multirow{2}{*}{AP} & \multirow{2}{*}{AP\textsubscript{50}}  & \multirow{2}{*}{AP\textsubscript{75}} & \multirow{2}{*}{AP\textsubscript{S}} & \multirow{2}{*}{AP\textsubscript{M}} & \multirow{2}{*}{AP\textsubscript{L}} &  \multicolumn{5}{c}{Sensitivity} \\ 
			\cmidrule{8-12}
			&&&&&&& 0.5 & 1 & 2 & 4 & mFROC \\
			\midrule 
			Faster RCNN \cite{b8} & 30.6 & 46.2 & 32.8 & 9.5 & 43.4 & 34.3 & 60.8 & 68.6 & 68.6 & 68.6 & 66.6 \\
			Cascade RCNN \cite{b12} & 31.4 & 45.7 & 33.4 & 10.4 & 44.7 & 36.4 & 66.2 & 67.6 & 67.6 & 67.6 & 67.2\\
			RetinaNet \cite{b15} & 19.2 & 32.4 & 20.8 & 10.4 & 26.9 & 31.9 & 59.7 & 67.9 & 72.5 & 75.2 & 68.8\\
			CenterNet \cite{b16} & 32.0 & 48.5 & 35.6 & 10.0 & 44.4 & 35.6 & 67.1 & 72.9 & 80.7 & 81.8 &  75.6\\			
			TOOD  \cite{b18} & 32.7 & 46.6 & 35.3 & 16.8 & 44.1 & 38.8 & 66.7 & 79.5 & 80.6 & 82.2 & 77.2\\
			Sparse RCNN \cite{b19} & 24.3 & 42.5 & 25.0 & 4.8 & 33.4 & 32.9 & 59.9 & 62.8 & 65.4 & 66.5 & 63.6 \\
			YOLOx  \cite{b20} & 35.8 & 48.2 & 41.2 & 11.5 & 45.8 & 42.6 & 60.3 & 65.5 & 65.8 & 65.8 &  64.3\\
			YOLOv8 \cite{b21} & 37.0 & 49.1 & 41.7 & 19.0 & 42.4 & 38.8 & 61.2 & 63.5 & 66.7 & 66.7 & 64.5 \\
			Deformable DETR  \cite{b22} & 10.6 & 20.7 & 11.4 & 2.8 & 16.4 & 6.9 & 23.5 & 34. 5 & 36.3 & 36.3 & 32.6\\
			DINO \cite{b23} & 37.1 & 52.5 & 40.5 & 14.0 & 45.9 & \textbf{45.1} & 67.3 & 71.2 & 73.9 & 77.2 & 72.4\\
			\midrule 
			FCOS(Baseline) \cite{b17} & 36.1 & 55.8 & 39.0 & 15.4 & 44.8 & 41.0 & 73.7 & 78.6 & 80.8 & \textbf{82.9} & 79.0\\
			Our Method & \textbf{42.0} & \textbf{62.2} & \textbf{46.7} & \textbf{22.3} & \textbf{50.0} & 44.1 & \textbf{75.3} & \textbf{80.0} & \textbf{80.9} & 82.3 & \textbf{79.5}\\
			\bottomrule 
		\end{tabular}
		\label{tab1}
	\end{center}
\end{table*}

\subsection{Dataset}
We conducted experiments on two datasets. The first pancreatic lesion dataset is a contrast-enhanced CT images of pancreas provided by the First Affiliated Hospital, Zhejiang University School of Medicine. The dataset contains 1482 CT images, including 861 pancreatic serous cystic neoplasms (SCNs), 353 mucinous cystic neoplasms (MCNs) and 268 without tumors. CT slices of difficult-to-detect lesion (e.g., small lesions) comprise approximately 16\% of the dataset. 1173 images from the dataset are used for training, and other additional 309 images are allocated for testing. 

To validate the generalization performance of the model, we conducted experiments on the publicly available DeepLesion dataset \cite{b34}. This dataset contains 32,735 lesions on 32,120 axial slices from 4,427 patients. In this dataset, the lesion types are diverse and contain lesions from various sites (e.g., bone, abdomen, liver, etc.). The dataset is divided into training (70\%), validation (15\%), and test sets (15\%) following official standards. 

Data pre-process. Different HU window level and window width are set for each of the two datasets. Under different HU window, we can focus on lesions in certain specific organs. We set HU restriction in accordance with the window level and window width provided by expert radiologists, the pancreas dataset set 30 and 300, the Deeplesion dataset is set up as provided in the official documentation.
\subsection{Experimental Setting}
\textbf{Implementation Details.} All experiments are conducted on NVIDIA GeForce RTX 2080 11 GB GPUs. The input images training size is ${640 \times 640}$. The optimizer is set to stochastic gradient descent (SGD) with weight decay of 0.0001 and momentum of 0.9. The initial learning rate in our model is set to 0.02, which would automatically scaled according to batch size and GPU, and a total of 12 epochs are trained. Other comparative experiments retained the original design.

\textbf{Evaluation Metrics.} We mainly use two evaluation metrics: commonly used the Average Precision (AP) metric for object detection and the Free-Response Receiver Operating Characteristic (FROC). AP is defined as the area under precision-recall (PR) curve of a certain class, including AP,  AP\textsubscript{50}, AP\textsubscript{75}, AP\textsubscript{S}, AP\textsubscript{M}, and AP\textsubscript{L}. Mean Average Precision (mAP) refers to the average of the summed APs for each class. In this experiment, AP is used to represent the mAP result. The alternative evaluation metric, FROC is generally used in the medical field and allows the evaluation of arbitrary abnormalities on each image. Specifically, the detection of medical images requires extremely high sensitivity to ensure the detection of all abnormalities, allowing for some degree of false positives (FPs). FROC measures whether a detector can find more true positives (TPs) at the same false positive rate. In our experiments, we set the FPs per image to 0.5, 1, 2, and 4 to compare the sensitivity of different methods.

\subsection{Results}
In this section, we first evaluate the performance of our method and other different types of methods for detecting lesions on two datasets. Then ablation experiments are performed on the pancreas dataset.

\textbf{Comparison study.} For the pancreatic dataset, as shown in Table \ref{tab1}, our method shows improvements in both AP and sensitivity compared to the baseline. The AP has increased by approximately 6\%, and AP\textsubscript{S} has shown an improvement of around 7\%. In addition, under the more stringent mean FROC (mFROC) evaluation metric, our method improved by 0.5\%. Notably, at FPs in per images of 0.5, our method obtained a higher sensitivity, achieving an improvement of 1.6\%. 

Comparing our method with other two-stage, one-stage and transformer detectors, our method outperforms the others in overall performance,  as shown in Table \ref{tab1}. The CenterNet and TOOD show relatively good sensitivity under the FROC metric, it still falls short of our method's performance. Additionally, the Dino achieved performance similar to the baseline, yet in comparison to our method, only has a slightly higher AP\textsubscript{L} of 1\%, while all other metrics are inferior. 

\begin{table}[htbp]
	\caption{Comparison of detection performance of different algorithms on DeepLesion dataset. Using the AP evaluation metrics. "-" indicates that the indicator was not provided.}
	\begin{center}
		\begin{tabular}{ccccccc}
			\toprule 
			Method & AP & AP\textsubscript{50} & AP\textsubscript{75} & AP\textsubscript{S} & AP\textsubscript{M} & AP\textsubscript{L}  \\ 
			\midrule 
			Faster RCNN & 29.1  & 52.1  & 30.7 & 0.1 & 34.7 & 44.6  \\
			Cascade RCNN & 29.8  & 51.7 & 32.6 & 0.1 & 35.7 & 45.4  \\
			RetinaNet & 25.4 & 48.4 & 23.8 & 0.1 &  29.0 & 47.9  \\
			CenterNet & 32.0 & 57.3 & 33.6 & 0.1 & 38.3 & 49.8 \\			
			TOOD & 33.5 & 59.3 & 35.6 & 0.3 & \textbf{41.6} & 50.8\\
			Sparse RCNN & 26.2 & 48.7 & 25.6 & 0.2 & 33.7 & 45.7  \\
			YOLOx & 32.2 & 58.0 & 33.3 & 0.1 & 37.5 & 49.7  \\
			Deformable DETR & 24.5 & 47.7 & 22.4 & 0.2 & 29.5 & 39.5 \\
			DINO & 33.3 & 57.4 & \textbf{36.1} & 0.2 & 40.3 & 47.9 \\
			Liu \cite{b29} & -  & 57.5 & - & - & - & - \\
			Zhu \cite{bmm} & -  & 60.4 & - & - & - & - \\
			\midrule 
			FCOS(Baseline) & 32.3 & 58.4 & 32.9 & 0.1 & 37.7 & 51.5  \\
			Our Method & \textbf{33.5} & \textbf{60.7} & 34.7 & \textbf{0.5} & 38.8 & \textbf{52.0} \\
			\bottomrule 
		\end{tabular}
		\label{tab2}
	\end{center}
\end{table}

For the Deeplesion dataset, which contains lesion types from different organ sites, it is more complex to realize the detection task compared to the pancreas dataset. As indicated in Table \ref{tab2}, our method also achieves superior results on the Deeplesion dataset, with an AP\textsubscript{50} reaching 60.0\%, and it improves the AP\textsubscript{S} from 0.1\% to 0.5\%, demonstrating the generalization of the model. In this dataset, the proportion of small lesions is extremely low. This scarcity of small lesions makes it more challenging to learn features related to small targets during the training process, leading to difficulties in detecting these small lesions. 

\begin{figure}[htb]
	\centering
	\includegraphics[width=0.98\linewidth]{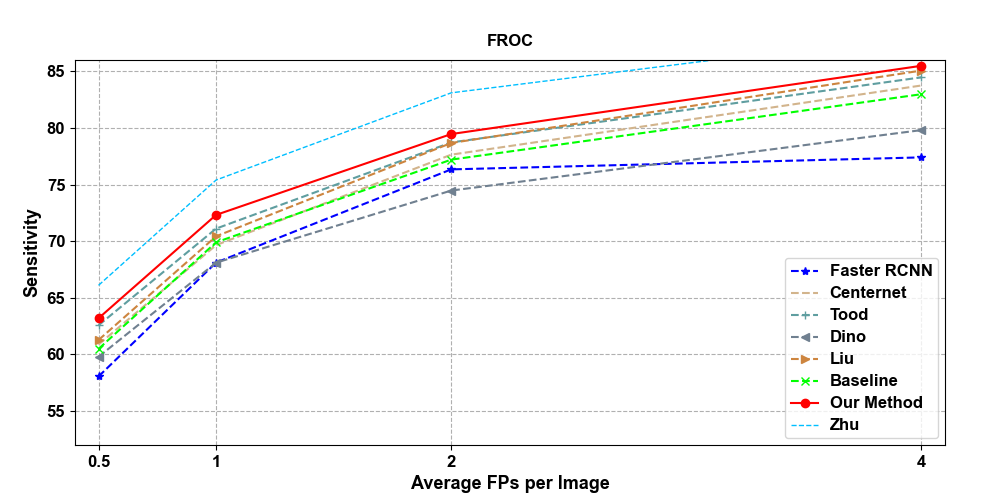}
	\caption{The FROC curves of various methods for detection on the Deeplesion dataset.}
	\label{figure 5}
\end{figure}

TOOD does not differ much from the performance of our method, but the AP\textsubscript{50}, AP\textsubscript{S} and AP\textsubscript{L} are 1.4\%, 0.2\% and 1.2\% higher than TOOD, respectively. when comparing more stringent FROC metrics, our sensitivity is higher the TOOD, as shown in Fig.~\ref{figure 5}. FROC curves visually compare the sensitivity of methods with similar AP values, and our method were higher than the baseline at average FPs per image = 0.5, 1, 2, and 4 by 2.7\%, 2.4\%, 2.2\%, and 2.5\%, respectively. However, compared to Zhu's method\cite{bmm}, our FROC metric still fails to fully outperform it, despite performing better on the AP50. We will continue to make improvements in subsequent studies to optimize our approach.

\begin{figure}[htb]
	\centering
	\includegraphics[width=0.9\linewidth]{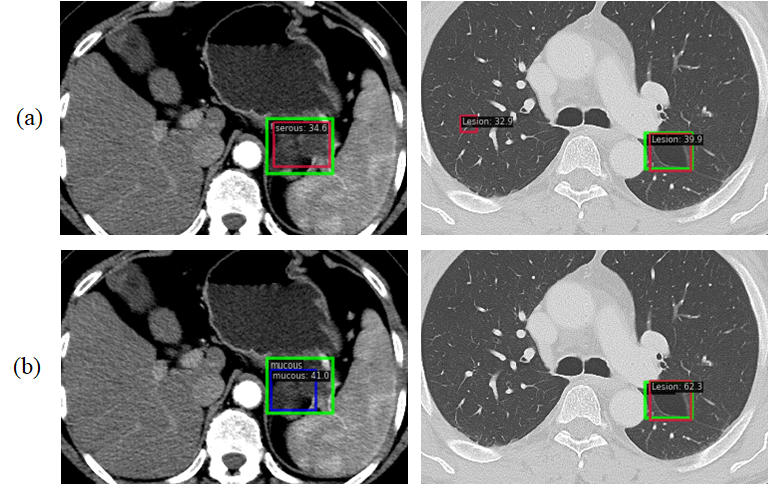}
	\caption{Detection visualization of the comparison between baseline and our improved detector. (a): the results of baseline. (b): the results of our method. The green and red bounding boxes represent  ground truths and predictions.}
	\label{figure 6}
\end{figure}

As shown in Fig.~\ref{figure 6}, the first column visualizes the detection results of the slices in the pancreas dataset and the second column shows the results of the Deeplesion dataset. Our method (b) is better at recognizing categories compared to baseline (a), since there is no misdiagnosis of mucous cyst as serous cyst. Furthermore, the baseline results may exhibit some false positives, while our method avoids such occurrences and achieves higher confidence. 

To validate the effectiveness of our designed ICAF-FPN structure, we replace the FPN structure in the neck part of baseline and compare the performance of detectors with different FPN mechanisms, as shown in Table \ref{tab3}. Comparing with FPN, our method improves the AP by 6\% and the AP\textsubscript{S} by 7\%, which is a significant improvement in the overall performance. Despite using the structure of PAFPN to improve the effectiveness, our method still outperformed PAFPN by 3\%. This demonstrates that our proposed approach effectively improves lesion detection performance.

\begin{table}[htbp]
	\caption{Ablation study on FPN, detection performance under different FPN structures.}
	\begin{center}
		\begin{tabular}{ccccccc}
			\toprule 
			Neck & AP  & AP\textsubscript{50} & AP\textsubscript{75} & AP\textsubscript{S} & AP\textsubscript{M} & AP\textsubscript{L} \\ 
			\midrule 
			FPN \cite{b10} & 36.1 & 55.8 & 39.0  & 15.4 & 44.8 & 41.0 \\
			PAFPN \cite{b24} & 39.4 & 59.7 & 42.1 & 17.9 & 47.4 & 42.7  \\
			NAS-FPN \cite{b25} & 24.3 & 37.7 & 25.9& 10.3 & 32.9 & 32.7  \\
			\midrule 
			ICAF-FPN & \textbf{42.0} & \textbf{62.2} & \textbf{46.7} & \textbf{22.3} & \textbf{50.0} & \textbf{44.1} \\
			\bottomrule 
		\end{tabular}
		\label{tab3}
	\end{center}
\end{table}

\textbf{Ablation study.} We perform ablation experiments on the proposed method to evaluate the performance improvement in lesion detection. The same strategy and parameters are used during training and validation to ensure a fair comparison.

We evaluated the impact of different designs in the ICAF-FPN on the model. As shown in Table \ref{tab4}, when only the ICA block is introduced, resulted in an AP of 39.0\%, an improvement of 2.9\% compared to the baseline. Notably, the AP\textsubscript{S} also achieves an improvement of 2.7\%. This indicates that combining background information can help with lesion detection.  When only introducing the AFW block for information interaction across different scales, we achieved better results for small target detection, with a 4.1\% improvement over the baseline, while other metrics also showed improvements. This reflects the fact that our method retains more details that contribute to lesion identification. When combining ICA and AFW and not using AFF, we noticed that AP\textsubscript{S} and AP\textsubscript{L} are not as high as when using AFW blocks alone. We conjecture that feature redundancy and conflicts occur when using additional fusion of different semantic features, leading to confusion between small and large target information. To address this, we introduce adaptive features fusion (AFF) mechanism by assigning different weights to the features. As shown in the Table \uppercase\expandafter{\romannumeral4}, after the introduction of AFF, AP\textsubscript{S} reached 22.3\%, and AP\textsubscript{L} goes from 41.6\% to 44.1\%, which is an improvement in each metric, proving the effectiveness of the adaptive module.
\begin{table}[htbp]
	\caption{Ablation study on ICAF-FPN, comparison of the performance for different blocks. }
	\begin{center}
		\begin{tabular}{cccccccc}
			\toprule 
			ICA & AFW  & AFF & AP & AP\textsubscript{50} & AP\textsubscript{S} & AP\textsubscript{M} & AP\textsubscript{L}\\ 
			\midrule 
			& & &36.1 & 55.8 & 15.4 & 44.8 & 41.0  \\
			\checkmark& & & 39.0 & 58.0 & 18.1 & 46.0 & 43.4 \\
			& \checkmark & & 39.1 & 58.1 & 19.5 & 45.1 & \textbf{44.6}    \\
			\checkmark& \checkmark & & 39.6 & 60.9 & 18.9 & 46.9 & 41.6  \\
			\checkmark& \checkmark & \checkmark & \textbf{42.0} & \textbf{62.2} & \textbf{22.3} & \textbf{50.0} & 44.1  \\
			\bottomrule 
		\end{tabular}
		\label{tab4}
	\end{center}
\end{table}

In the AFW block, to retain more information about small targets, we additionally introduced the C2 layer compared to the baseline. To verify the necessity of the C2 layer, we conducted an ablation experiment, as shown in Table \ref{tab5}. 

\begin{table}[htbp]
	\caption{Ablation study on AFW block, w/o indicates that no C2 layer was introduced, w/ indicates that a c2 layer was introduced.}
	\begin{center}
		\begin{tabular}{cccccccc}
			\toprule 
			ICAF-FPN & AP & AP{50} & AP\textsubscript{75} & AP\textsubscript{S} & AP\textsubscript{M} & AP\textsubscript{L}\\ 
			\midrule 
			w/o C2 & 38.0 & 57.3 & 41.8 & 14.2 & 46.8 & \textbf{44.1}  \\
			w/  C2 & \textbf{42.0} & \textbf{62.2} & \textbf{46.7} & \textbf{22.3} & \textbf{50.0} & \textbf{44.1} \\
			
			\bottomrule 
		\end{tabular}
		\label{tab5}
	\end{center}
\end{table}

Our method fully utilizes the information extracted from both inter- and intra-layers to enhance the overall detection performance, which also effectively improving the detection rate of small targets without losing the accuracy of large targets.

\section{Conclusion}
In this paper, we propose a lesion detector based on one-stage method called IAFI-FCOS for assisting radiologists to achieve faster and more accurate early screening and detection of cancer lesions. We mainly focus on enhancing the neck part of the object detection framework, which extracts lesion-related information from both intra- and across-scale perspectives for features of different scales. Subsequently, we adaptively fuses the diverse semantic features of each scale and feed into the prediction head to obtain the final detection results. The novel multi-scale feature fusion mechanism ICAF-FPN alleviates the challenge of detecting ambiguous and small lesions. Extensive experiments have demonstrated a significant improvement in the detection of the pancreatic lesion dataset, as well as an enhanced detection performance of the Deeplesion dataset. Continuous efforts are still needed to achieve full generalizability across various lesion detection domains. In future studies, we will continue to improve and optimize our method to enhance domain generalizability and better meet the needs of the medical image analysis field.

\end{document}